\pdfoutput=1

\documentclass{article}
\usepackage{ijcai17}

\usepackage{times}
\usepackage{tabularx}
\usepackage{booktabs}
\usepackage{threeparttable}
\usepackage{algorithm}
\usepackage{algorithmic}
\usepackage{amsmath}
\usepackage{multirow}
\usepackage{graphicx}
\usepackage{caption}

\usepackage[encapsulated]{CJK}
\usepackage{ucs}
\usepackage[utf8x]{inputenc}
\title{Learning to Identify Ambiguous and Misleading News Headlines}
\author{Wei Wei and Xiaojun Wan\\
Institute of Computer Science and Technology, Peking University\\
The MOE Key Laboratory of Computational Linguistics, Peking University  \\
\{weiwei718,wanxiaojun\}@pku.edu.cn}

\begin{document}

\maketitle

\begin{abstract}
Accuracy is one of the basic principles of journalism. However, it is increasingly hard to manage due to the diversity of news media. Some editors of online news tend to use catchy headlines which trick readers into clicking. These headlines are either ambiguous or misleading, degrading the reading experience of the audience. Thus, identifying inaccurate news headlines is a task worth studying. Previous work names these headlines ``clickbaits'' and mainly focus on the features extracted from the headlines, which limits the performance since the consistency between headlines and news bodies is underappreciated. In this paper, we clearly redefine the problem and identify ambiguous and misleading headlines separately. We utilize class sequential rules to exploit structure information when detecting ambiguous headlines. For the identification of misleading headlines, we extract features based on the congruence between headlines and bodies. To make use of the large unlabeled data set, we apply a co-training method and gain an increase in performance. The experiment results show the effectiveness of our methods. Then we use our classifiers to detect inaccurate headlines crawled from different sources and conduct a data analysis.
\end{abstract}

\section{Introduction}

With the rapid development of the Internet, a variety of online news websites spring up and attract a great number of readers by providing much convenience and a wealth of information. However, the prosperity of online journalism simultaneously brings about problems including inaccuracy, unfairness, and subjectivity. We are to solve the problem of inaccurate news headlines since they are frequently complained about by readers.

With the purpose of attracting clicks, online news publishers use diverse strategies to make their headlines catchy. Generally, those clickbait headlines can be classified into two categories, namely ambiguous ones and misleading ones \cite{marquez:1980accurate}. Ambiguous headlines make use of curiosity gap by concealing key information of a news event. As a result, readers tend to click on the links and find out the missing information. Misleading headlines always exaggerate or distort the fact described in news bodies. Readers will discover inconsistencies after reading the whole passage, only to be left disappointed.

News with inaccurate headlines is unacceptable. It is more than a violation of professional ethics for editors. From the perspective of users, the content of this kind of news hardly lives up to their expectations comparing with the catchy headlines. Readers usually decide whether it is worth their time to read a story through headlines, so tricky headlines will waste their time and degrade the user experience. More importantly, it has been proved that headlines can shape public opinion~\cite{tannenbaum1953effect}. According to the study, readers think differently of the accused after reading news with headlines slanted toward either innocence or guilt. From the above, we can see that it is necessary to solve the problem. Hence, we turn to the study of automatically identifying ambiguous and misleading headlines.

In this paper, we redefine the problem using Journalism and Communication knowledge. Instead of simply treating each piece of news as clickbait or non-clickbait, we separately judge whether it is ambiguous and whether it is misleading. For the ambiguous headline detection task, we mine class sequential rules (CSR) to make use of sequential information. Previous work mostly employs features extracted from headlines~\cite{chakraborty2016stop,anand2016we}. Biyani\emph{ et al.}~\shortcite{biyani20168} utilized the informality of news body, but the consistency is still underappreciated. In our method for misleading headline detection, we extract features based on headlines, bodies and the relationship between them. Since we only annotate a small part of data set, we also design a semi-supervised method, co-training, to exploit the unlabeled news. Body-independent and body-dependent features are utilized to train the sub-classifiers.

The experiment results show the effectiveness of CSR features for the ambiguous headline detection task and the consistency features for the misleading headline detection task, and performance improves when co-training is used. With the final classifiers, we classify news crawled from four major Chinese news websites, which cover different categories such as sports, society and world news. Then we conduct a data analysis upon the results.

\section{Related Work}

In Communication and Psychology, there have been studies on the accuracy of news headlines since decades ago. Marquez~\shortcite{marquez:1980accurate} divided news headlines into three types, namely accurate, ambiguous and misleading ones, and proposed specific definitions, which are subsequently used in our classification. Ecker \emph{et al.}~\shortcite{ecker:2014effects} analysed the effect of misinformation in news headlines and showed that such headlines lead to misconception in readers. Several properties and structures of clickbait headlines were dug out manually~\cite{molek2013towards,molek2014coercive,blom2015click}, providing an entry point for preliminary automatic identification of clickbaits.

It was not until recent years that some studies on clickbait detection emerged in the field of artificial intelligence. Chakraborty \emph{et al.}~\shortcite{chakraborty2016stop} extracted a set of features from headlines to train a clickbait classifier. Instead of using hand-crafted features, Anand \emph{et al.}~\shortcite{anand2016we} tried RNN method with word embeddings as inputs. Some potential non-text cues, such as user behavior analysis and image analysis, were discussed but not implemented by Chen \emph{et al.}~\shortcite{chen2015misleading}. Biyani \emph{et al.}~\shortcite{biyani20168} further utilized both title-based and body-based (article informality) features to identify clickbait news. However, all the above works hardly considered the relationship between headlines and bodies, and focused more on the properties of headlines. In this paper, we clearly define the problem with Marquez's professional view~\shortcite{marquez:1980accurate} rather than the general wording ``clickbait'', detecting ambiguous headlines and misleading ones separately. In the latter task, the consistency between headlines and bodies is especially taken into account.

\section{Problem Definition and Corpus}

\subsection{Problem Definition}

From the perspective of Journalism and Communication, news headlines are classified into the three categories below~\cite{marquez:1980accurate}. Note that a headline can be both ambiguous and misleading. In this paper, we divide our task into two separate parts: ambiguous headline detection and misleading headline detection.

\subsubsection{Accurate Headlines}

\emph{An accurate headline is a headline that is congruent in meaning with the content of the news story.}

Such headlines are non-clickbaits.

\subsubsection{Ambiguous Headlines}
\label{ambeg}
\emph{An ambiguous headline is a headline whose meaning is unclear relative to that of the content of the story.}

It is typical of ambiguous headlines to omit some key information. The lack of knowledge arouses reader's curiosity and lures them to click. For example:

\begin{CJK*}{UTF8}{gbsn}
\small “她是曾经的世界冠军，但现在为工作发愁。”
\end{CJK*}

(``She once won the world championships, but now worries about making a living.'')

\subsubsection{Misleading Headlines}

\emph{A misleading headline is a headline whose meaning differs from that of the content of the story.}

The differences can be either subtle or obvious. Some common tactics are exaggeration, distortion, etc, most of which aim to cause a sensation. For example:

\begin{CJK*}{UTF8}{gbsn}
\small “亚洲鲤鱼已经成为美国的噩梦。”
\end{CJK*}

(``Asian carp has become the nightmare of America!'')

In the news body of this example, the overgrowth of Asian carp due to the lack of natural predators is described. The word ``nightmare'' and the exclamation mark actually exaggerate the problem.

\subsection{Corpus}
\label{sec:corpus}
To cover a wide range of news, we crawled a total of 40 000 articles in six different domains (domestic, world, society, entertainment, sports, and technology) from four major Chinese news sites (Sina, NetEase, Tencent, and Toutiao).

Because of the exact definition of the problem, annotators have to read through the news headlines and bodies, which is a time-consuming and demanding job. Therefore, we randomly select 2924 pieces of news and employ 6 college students majoring in Chinese to label each news headline as ambiguous or not, and also label it as misleading or not. They have read relevant instructions before annotating and each piece of news is labeled by at least 3 people. To ensure the consistency of misleading headline annotation, we abandoned 316 controversial examples (at least one annotation differs from the others). The final labeled data set contains: 645 ambiguous and 2279 non-ambiguous; 843 misleading and 1765 non-misleading.

Note that 24 000 pieces of unlabeled news are used for co-training, and a big data analysis is conducted upon the full data set. Some of the articles are not used for co-training because we continuously crawl them after the experiments.

\section{Identifying Ambiguous Headlines}

According to the definition of ambiguous news headlines, they usually deliberately omit some main elements of sentences to spur curiosity, which can be visually seen without reading news bodies. Thus, we firstly follow the previous practice to extract features from headlines. But the downside is that those features are mainly word-based, losing sight of sentence structures and sequential information. Thus, we secondly mine class sequential rules (CSR) and then derive CSR features. Finally, both the basic features and CSR features are utilized to train an SVM classifier~\cite{joachims2002learning}. SVM machine learning method is selected in this task for its outperformance over other methods. We use the SVM toolkit in scikit-learn\footnote{http://scikit-learn.org/stable/modules/generated/sklearn.svm.SVC.html}.

\subsection{Basic Features}

Table 1 lists the basic features. Those features were proved to be effective in English clickbait detection tasks~\cite{chakraborty2016stop}. Clickbait words are phrases or words commonly used in catchy headlines, such as ``You Won't Believe'' and ``Will Blow Your Mind''. We translate the English clickbait word list released by Downworthy\footnote{http://downworthy.snipe.net/}, and manually extend the vocabulary to adapt the characteristics of Chinese. We make use of ``Chinese/English Vocabulary for Sentiment Analysis'' released by Hownet\footnote{http://www.keenage.com/} when counting sentiment words. This vocabulary contains six files, including words expressing sentiment degree, subjective opinion, positive evaluation, negative evaluation, positive emotion, and negative emotion. We match internet slang using the lexicons released by SogouInput\footnote{http://pinyin.sogou.com/dict/}.
\begin{CJK*}{UTF8}{gbsn}

\begin{table}
\small
\centering
\begin{threeparttable}
\begin{tabularx}{3.375in}{lX}
\toprule
Feature   & Description  \\
\midrule
Wordcnt   & Count the number of words \\
Number   & Count the number of numerals \\
Baitword   & Count the number of clickbait words \\
Slang   & Count the number of internet slang \\
Punctuation   & Count the number of !, ? and ... \\
SentDegree   &  Respectively count 2 sets of degree adverbs expressing “很” (very) and “非常” (extremely) \\
SentPolar   & Respectively count words expressing positive evaluation, negative evaluation, positive emotion, and negative emotion \\
Distance   & Compute the average distance between governing and dependent words (identified by LTP parser\footnotemark)\\
WHword   & Count the number of Chinese interrogative pronouns\\
ForwardRef   & Count words expressing forward-reference, including demonstratives (this,that,...) and personal pronouns (he,she,it,...)\\
\bottomrule
\end{tabularx}
\renewcommand{\captionfont}{\small}
\caption{Basic features extracted from headlines}
\label{table:feat}
\end{threeparttable}
\end{table}
\footnotetext{http://ltp.readthedocs.io/}

\end{CJK*}
\subsection{Class Sequential Rules Mining}

We utilize a sequential pattern mining method described in ~\cite{liu2007web} to find language patterns of ambiguous headlines and non-ambiguous ones, and then derive features based on the sequential rules.

Let $I = \{i_{1}, i_{2},..., i_{n}\}$ be a set of items. A sequence $s$ is an ordered list of items. $s$ is denoted by $\left \langle a_{1},..., a_{i},..., a_{r}\right \rangle$, where $a_{i}$ is an item in $I$. A sequence $s_{1} = \left \langle a_{1}, a_{2},...,a_{n}\right \rangle$ is called a subsequence of $s_{2} = \left \langle b_{1}, b_{2},...,b_{m}\right \rangle$ if there exist integers $1 \le j_{1} \le j_{2} \le ... \le j_{n-1} \le j_{n}$ such that $a_{1} = b_{j_{1}}, a_{2} = b_{j_{2}}, ..., a_{1} = b_{j_{n}}$.

The sequence database $D$ is a set of pairs $\left\{(s_{1}, y_{1}), (s_{2}, y_{2}),..., (s_{n}, y_{n})\right\}$, where $s_{i}$ is a sequence and $y_{i}\in Y$ is a class label. In the context of ambiguous headlines detecting, $Y = \left\{ambiguous, non\text{-}ambiguous\right\}$. A class sequential rule (CSR) is of the following form:
\begin{equation*}
X \rightarrow y \text{, where $X$ is a sequence and $y \in Y$}
\end{equation*}

An instance $(s_{i}, y_{i})$ \textbf{covers} the CSR if $X$ is a subsequence of $s_{i}$. An instance $(s_{i}, y_{i})$ \textbf{satisfies} the CSR if $X$ is a subsequence of $s_{i}$ and $y_{i} = y$. The \textbf{support} (sup) of a CSR is the fraction of instances in D that satisfy the rule. The \textbf{confidence} (conf) of the rule is the proportion of instances that cover the rule which also satisfy the rule.

To explain the above definitions, we consider the following example shown in Table \ref{table:csr}. Setting the minimum support to be 0.2 and the minimum confidence to be 0.4, one of the CSRs is $\left \langle 1, 4, 7\right \rangle \rightarrow c_{1}$. The support of this rule is 0.4 since sequence 1 and 2 satisfy the rule, and the confidence is 0.67 since sequence 1, 2 and 5 cover the rule.

\begin{table}[h!]
\centering
\small
\begin{tabular}{|p{1cm}<{\centering}|p{2.5cm}<{\centering}|p{1cm}<{\centering}|}
\hline
    ID      & Sequences & Class  \\
\hline
1 & $\left \langle 1, 4, 5, 6, 7\right \rangle$ & $c_{1}$  \\
\hline
2 & $\left \langle 1, 4, 6, 7, 9\right \rangle$ & $c_{1}$  \\
\hline
3 & $\left \langle 1, 6, 7\right \rangle$ & $c_{1}$  \\
\hline
4 & $\left \langle 2, 6, 7\right \rangle$ & $c_{2}$  \\
\hline
5 & $\left \langle 1, 3, 4, 7\right \rangle$ & $c_{2}$  \\
\hline
\end{tabular}
\renewcommand{\captionfont}{\small}
\caption{An example database of CSR mining}
\label{table:csr}
\end{table}

To build a sequence database, we transform the news headlines into sequences. The item set $I$ contains 12 labels corresponding to word types mentioned in Table \ref{table:feat} such as WH-words and forward-reference words. $I$ also contains 2 labels of temporal adverbs expressing ``past'' and ``present'', as well as 9 labels corresponding to different Chinese conjunctions. Temporal adverbs are used here because some headlines cause a sensation by presenting a contrast between the past and present. In Chinese, conjunctions may increase the attractiveness of a sentence by indicating antithesis, hypothesis, etc. For each word in a headline, it may be transformed into a type label. As a result, a headline, which is originally a sequence of words, is encoded into a sequence of type labels. If there is no match between a word with any items in the vocabularies, we just skip it and move on.

Take the sentence in Section \ref{ambeg} as an example. We search for each word in the lexicons, and the word\begin{CJK*}{UTF8}{gbsn} “她” (she), “曾经” (once), “但”(but) and “现在” (now)\end{CJK*} are successfully matched. Consequently, the word sequence is transformed into a label sequence as follows:
\begin{equation*}
\left \langle Ref, Past, But, Present\right \rangle
\end{equation*}
With selected minimum support (minsup) and minimum confidence (minconf), we utilize the CSR mining algorithm described in ~\cite{liu2007web} to mine CSRs from the training database. Then the sequential pattern $X$ in each CSR is treated as a feature, which is set to 1 when being contained by a headline.

After extracting features based on CSRs, we conduct experiments with CSR features to demonstrate the efficacy. We also train classifiers with all features mentioned above and see a performance improvement comparing with basic methods.

\section{Identifying Misleading Headlines}

A headline is considered misleading if and only if it differs from the news body. Hence, features extracted separately from headlines and bodies cannot provide enough evidence for this task. We select four groups of features which evaluate the consistency between headlines and bodies. Additionally, we utilize a bootstrapping method, co-training, to take advantage of the large set of unlabeled data.

\subsection{Features}

\subsubsection{Body-independent Features}

This set of features is extracted from headlines only, similar to those included in Table \ref{table:feat}. Note that $ForwardRef$ is abandoned here since this feature merely reflects ambiguity.

\subsubsection{Body-dependent Features}

Some of the following features are derived from news bodies, while others reflect the consistency between headlines and bodies.

\begin{enumerate}
  \item \textbf{Informality:} We compute the frequencies of two informality indicators, namely internet slang and bait words. Additionally, the length of news bodies is also an input feature.
  \item \textbf{Sentiment:} Sentiment feature consists of the frequencies of positive evaluations, negative evaluations, positive emotions, negative emotions and subjective words.
  \item \textbf{InformalGap: } We calculate the absolute value of informality difference between headlines and bodies of each piece of news.
  \item \textbf{SentiGap:} We calculate the absolute value of sentiment difference between headlines and bodies of each piece of news. Specifically, this feature set contains the differences in five frequencies mentioned in $Sentiment$.
  \item \textbf{Similarity:} Misleading headlines often contains words that differ from those in news bodies. Thus, we firstly count the number of named entities that occur in a headline but absent in the corresponding news body. Secondly, for each word (except entities) $h_{i}$ in a headline, through calculating the cosine distance of word embeddings, we find its most similar word $b$ in body, and record this largest cosine similarity as $s_{i}$. Then the following values are used as features:
      {\setlength\abovedisplayskip{1.5pt}
       \setlength\belowdisplayskip{1pt}
      \begin{equation*}
      \small
      minSim = \min_{i=1}^{m}s_{i}
      \end{equation*}}
      {\setlength\abovedisplayskip{1pt}
       \setlength\belowdisplayskip{1pt}
      \begin{equation*}
      \small
      avgSim = \sum_{i=1}^{m}s_{i}/m
      \end{equation*}}
      where $m$ is the number of words except named entities in a headline. Thirdly, tf-idf is used to compute the overall similarity between a headline and a news summary generated by PKUSUMSUM~\cite{zhangpkusumsum}.
  \item \textbf{Recognizing Textual Entailment(RTE):} Textual entailment is defined as follows: a text T entails another text H if the meaning of H can be inferred from the meaning of T with common background knowledge. Previous studies proposed RTE methods based on the similarity of dependency trees~\cite{kouylekov2005recognizing,wang2007recognizing}. To simplify the problem, we parse the headline and sentences in body into dependency trees, and calculate RTE-score by matching pairs of governing and dependent words which are sentence skeletons. Specifically, we search sentences in the body for dependency pairs occurred in the headline. While matching, synonym, hypernym, hyponym, and antonym are taken into account and assigned different weights. RTE-score equals the weighted sum of each match.
\end{enumerate}

\subsection{The Co-Training Approach}

Identifying misleading headlines is a relatively complicated problem due to the variety of baiting strategies, while we only have a small set of labeled data. In order to build a robust classifier that is capable of handling different instances, we try to make full use of the larger unlabeled data set via a semi-supervised method.

Co-training~\cite{blum1998combining} is a typical semi-supervised method, which demands features based on two independent views, but the independence assumption can be relaxed. Starting with the limited data set, co-training can increase the amount of labeled data by annotating the unlabeled data with two sub-classifiers. In recent years, co-training has been successfully applied to co-reference resolution~\cite{ng2003weakly}, sentiment classification~\cite{wan2009co}, review spam identification~\cite{li2011learning}, etc.

Since we extract body-independent and body-dependent features from each piece of news, our task exactly satisfies the essential requirement of co-training. The algorithm framework is shown in Algorithm \ref{alg:1}. In the experiments, we balance the parameter values of $p$ and $n$ at each iteration to maintain the class distribution in the labeled data. Through adding confidently predicted instances, two sub-classifiers gain useful information with the help of each other. Note that the examples with conflicting labels are excluded from $N_{h}$ $\cup$ $N_{b}$.

\begin{algorithm}
\caption{Co-Training Algorithm}
\label{alg:1}
\begin{algorithmic}[1]
\renewcommand{\algorithmicrequire}{ \textbf{Given:}}
\renewcommand{\algorithmicensure}{ \textbf{Loop for I iterations:}}
\REQUIRE Body-independent (headline-dependent) features $F_{h}$; body-dependent features $F_{b}$; a set of labeled news $L$; a set of unlabeled news $U$.
\ENSURE
\STATE Learn the first classifier $C_{h}$ from $L$ based on $F_{h}$;
\STATE Use $C_{h}$ to label news from $U$ based on $F_{h}$;
\STATE Choose $p$ positive and $n$ negative most confidently predicted news $N_{h}$ from $U$;
\STATE Learn the second classifier $C_{b}$ from $L$ based on $F_{b}$;
\STATE Use $C_{b}$ to label news from $U$ based on $F_{b}$;
\STATE Choose $p$ positive and $n$ negative most confidently predicted news $N_{b}$ from $U$;
\STATE Remove $N_{h}$ $\cup$ $N_{b}$ from $U$;
\STATE Add $N_{h}$ $\cup$ $N_{b}$ with the corresponding labels to $L$.

\end{algorithmic}
\end{algorithm}

During co-training, both the sub-classifiers will provide prediction scores for each instance. The prediction scores are normalized into $[0, 1]$. Finally, the average of the normalized values is used as the overall prediction score of each instance.

\section{Evaluation Results}
In this section, we set experiments to evaluate the performance of the two identification tasks. The corpus has been described in Section \ref{sec:corpus}. The evaluation metrics are \emph{precision}, \emph{recall}, and\emph{ F-score}.
%\begin{equation}
%precision= \frac{tp}{tp+fp}
%\end{equation}
%\begin{equation}
%recall= \frac{tp}{tp+fn}
%\end{equation}
%\begin{equation}
%F\text{-}score=\frac{2*precision*recall}{precision+recall}
%\end{equation}
%where $tp$, $fp$, and $fn$ refer to true positive, false positive, and false negative.
\subsection{Ambiguous Headlines Identification}
We randomly split the labeled data set into a training set and a test set in the proportion of 3:1. In the experiments, three versions of features are used for comparison:

\textbf{Unigram Features: }It is a baseline using the SVM classifier and the rbf kernel to identify ambiguous headlines, with unigram features provided. Unigrams here refer to Chinese words but not characters.

\textbf{Basic Features: }It uses the SVM classifier and the rbf kernel to identify ambiguous headlines, with only basic features provided.

\textbf{CSR Features: }It uses the SVM classifier and the rbf kernel with only CSR Features provided. When mining class sequential rules, we conduct experiments with different minimum support and minimum confidence. In Table 3, we list the result of $minsup$ 0.02 and $minconf$ 0.8 since this set of parameters fit well on the training set.

\textbf{All Features: }It uses the SVM classifier with both basic features and CSR features provided. The $minsup$ and $minconf$ for mining are the same as above.

\begin{table}[h!]
\centering
\small
\begin{tabular}{|l|c|c|c|}
\hline
          & Precision & Recall & F-score \\
\hline
Unigrams & 0.426 & 0.457 & 0.441 \\
\hline
CSR Features & 0.763 & 0.649 & 0.701 \\
\hline
Basic Features & 0.650 & 0.761 & 0.701 \\
\hline
All Features & 0.709 & 0.803 & 0.753 \\
\hline
\end{tabular}
\renewcommand{\captionfont}{\small}
\caption{Comparison results of methods identifying ambiguous headlines}
\label{table:ambres}
\end{table}

Table \ref{table:ambres} shows the comparison results. The method with unigram features does not perform well, because this task is different from traditional text classification tasks. The method with CSR features is proved to be effective and outperforms over basic features in precision, owing to its strictness with sentence structures. Our method with all features gets an increase in precision, recall, and F-score comparing with the method with basic features, demonstrating the contribution of CSR features. In addition, we conduct sign-test upon the predict results of basic features and all features. Sign test is a statistical method to test for consistent differences between pairs of observations. The $p$-$value$ of sign test is $0.0022 < 0.05 $, demonstrating the significant efficacy of our method.
\subsection{Misleading Headlines Identification}

In this task, the labeled data set is randomly split into a training set and a test set in the proportion of 3:1. We firstly conduct supervised learning to compare the efficacy of different features on the labeled data set. For co-training, the same test set is used for evaluation, and the training set as well as the unlabeled data is used for learning.

\subsubsection{Feature Validation}

To validate the effectiveness of each feature, we conduct supervised learning with all features and then exclude each group of body-independent features to compare the performance. In addition, we list the results of the method with all body-dependent features and the method with all body-independent features. We select the SVM classifier since it performs best comparing with other methods. Rbf kernel is used in the experiments. The baseline uses unigram features and the SVM classifier. The theoretical result of random classification is also provided for comparison, which is 0.323 because of the imbalance of the data set.

\begin{table}[h!]
\small
\centering
\begin{tabular}{|l|c|c|c|}
 \hline
          & Precision & Recall & F-score \\
 \hline
  Random & 0.323 & 0.323 & 0.323 \\
 \hline
  Unigrams & 0.428 & 0.456 & 0.443 \\
 \hline
  All Features(A) & 0.646 & 0.768 & 0.702  \\
 \hline
  Body-dependent & 0.602 & 0.660 & 0.630  \\
  Body-independent & 0.637 & 0.716 & 0.674  \\
 \hline
  A-Informality & 0.645 & 0.736 & 0.688  \\
  A-Sentiment & 0.646 & 0.752 & 0.695  \\
  A-Similarity & 0.645 & 0.756 & 0.696  \\
  A-RTEscore & 0.644 & 0.744 & 0.691 \\
  A-InformalGap & 0.640 & 0.742 & 0.687 \\
  A-SentiGap & 0.640 & 0.732 & 0.683 \\
 \hline
 \end{tabular}
 \renewcommand{\captionfont}{\small}
 \caption{Comparison results of different feature set}
 \label{tab:featvalid}
 \end{table}

Table \ref{tab:featvalid} shows the result of feature validation. With all features, the classifier achieves the F-score 0.702, which significantly exceed both sub-classifiers. According to the results listed, all the six groups of body-dependent features are resultful. Among all features, $InformalGap$ and $SentiGap$ are relatively important. Thus, it can be seen that the consistency between news headlines and news bodies plays an important role in misleading headline detection.

 \subsubsection{Co-Training Results}
In the previous section, the efficacy of two views of features is proved. In this section, we utilize the large number of unlabeled data by leveraging the co-training method. We compare the results of co-training and the supervised method to demonstrate the suitability of co-training in our task. The sub-classifiers use the same SVM-based machine learning method mentioned in the previous section. We use different sets of parameters and the experiment results listed in Table \ref{tab:co} is applied to the parameter set of $p=10, n=20$ and iteration number $=50$. Our proposed co-training method outperforms the supervised method with all features and gains an increase of 0.022 in F-score. We conduct sign test again and our method passes the significance test with a $p-value$ of $0.0183<0.05$.
\begin{table}[h]
\small
\centering
\begin{tabular}{|l|c|c|c|}
 \hline
          & Precision & Recall & F-score \\
 \hline
  Body-dependent & 0.602 & 0.660 & 0.630  \\
  Body-independent & 0.637 & 0.716 & 0.674  \\
 \hline
  All Features & 0.646 & 0.768 & 0.702  \\
 \hline
  Co-training & 0.670 & 0.788 & 0.724 \\
 \hline
 \end{tabular}
 \renewcommand{\captionfont}{\small}
 \caption{Comparison results of supervised method and co-training}
 \label{tab:co}
 \end{table}

\begin{figure}[h!]
  \centering\includegraphics[width=3in]{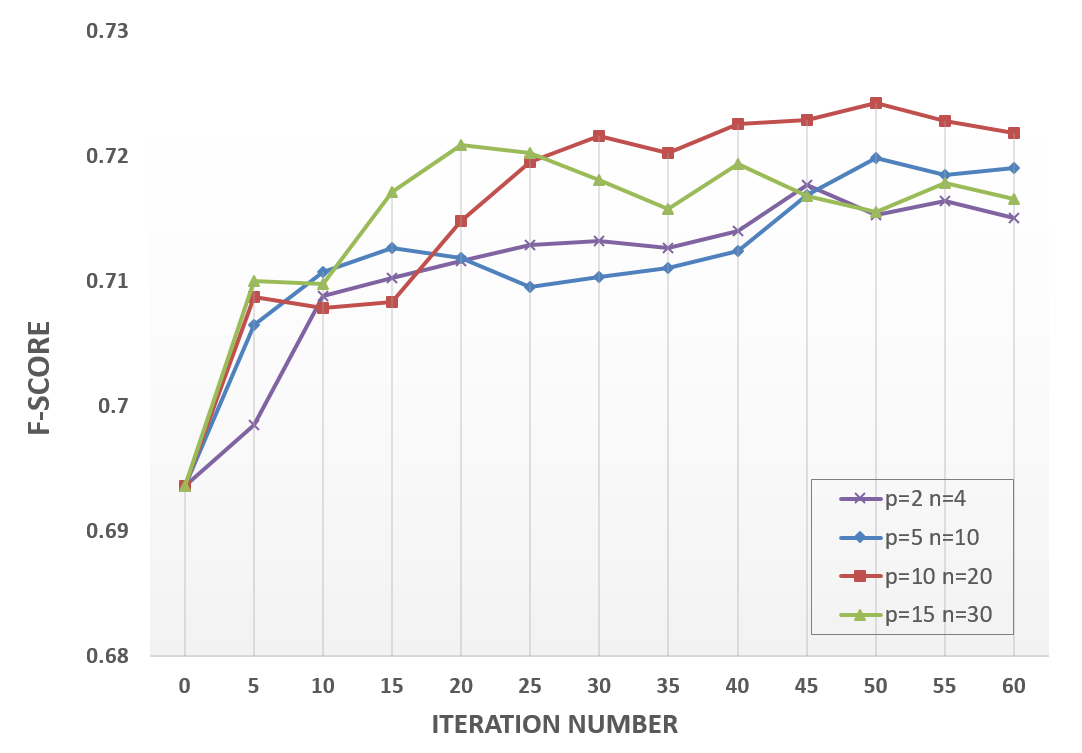}
  \renewcommand{\captionfont}{\small}
  \caption{Influence of parameters}
  \label{pic:co}
\end{figure}

\textbf{Parameter sensitivity:} For different parameter sets, the performance of our co-training method is slightly different. We change $p$ and $n$ in experiments, and also observe the results in different iteration numbers. Note that $n$ always equals $2p$ in order to keep the proportion of positive instances and negative instances. The results are shown in Figure \ref{pic:co}. We can see that at the beginning, F-score is on the rise along with the increasing iteration number. After 45 iterations, the results achieve relative stability with slight fluctuation. Among different growth sizes, $p=10$ and $n=20$ perform best on our data set. When $p=5$ or $2$, F-score increases slowly during iteration. For a larger value of $p=20$, the performance improves obviously in 20 iterations but no longer increase then, because a large growth size is less robust to noises.

\section{Data Analysis}

We used 40 000 pieces of news crawled evenly from four major Chinese news sites (Sina, NetEase, Tencent, and Toutiao). The data covers six domains including domestic, world, society, entertainment, sports, and technology. After training the two classifiers, we identify news with ambiguous headlines and misleading ones. Then we analyze the statistics, getting some interesting and useful results.

\begin{figure}[h!]
\centering\includegraphics[width=3in]{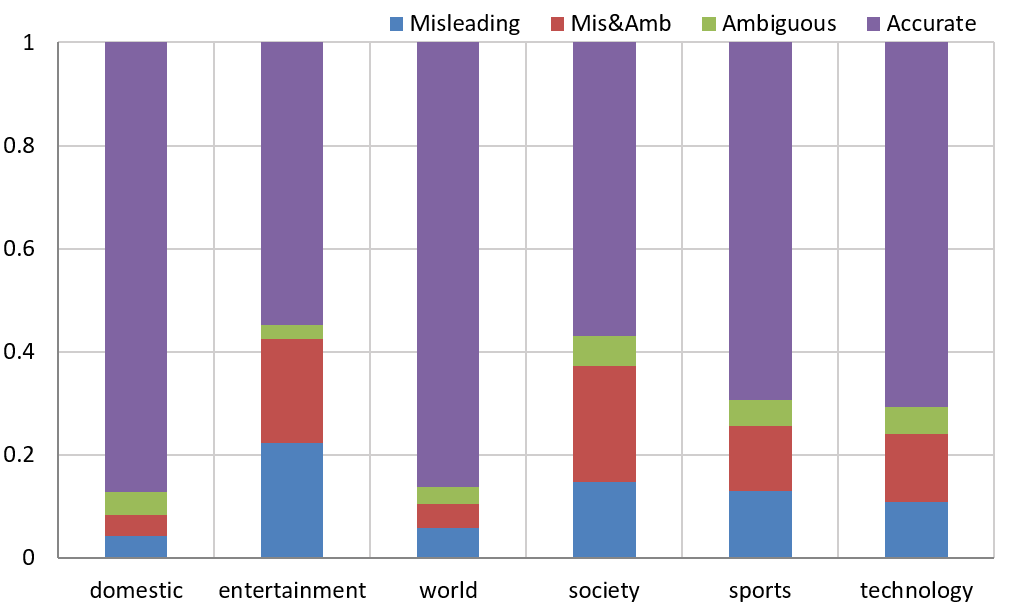}
\renewcommand{\captionfont}{\small}
  \caption{Comparison of different news domains}
  \label{pic:do}
\end{figure}

The comparison results among news in different domains are shown in Figure \ref{pic:do}. Domestic news and world news, which usually cover political and military events,  have a high rate of accuracy. The highest incidence of inaccurate headlines in entertainment news may result from its lack of seriousness. Society news telling unusual stories in social life also tends to have tricky headlines. In addition, this figure reflects that almost half the inaccurate headlines are both misleading and ambiguous.

\section{Conclusion}

In this paper, we study the problem of inaccurate headline detection. We crawl news online and build a Chinese data set. We divide the problem into two tasks. For ambiguous headline detection, we proposed a method based on class sequential rules and demonstrate the efficacy. For misleading headline detection, features evaluating the consistency between headlines and bodies are utilized. To exploit the larger unlabeled data set, we apply the co-training method in the latter task. The experimental results demonstrate the robustness of our identifiers.

Utilizing the final classifiers, we identify ambiguous and misleading headlines in the full data set. The results of data analysis reflect the need for more stringent regulations in journalism, especially for entertainment news and society news.

\section*{Acknowledgments}

This work was supported by 863 Program of China (2015AA015403), NSFC (61331011), and Key Laboratory of Science, Technology and Standard in Press Industry (Key Laboratory of Intelligent Press Media Technology). We thank the anonymous reviewers for helpful comments. Xiaojun Wan is the corresponding author.

\newpage
\bibliographystyle{named}
\bibliography{Clickbait_Detecting}
\nocite{marquez:1980accurate}
\nocite{molek2013towards}
\nocite{molek2014coercive}
\nocite{biyani20168}
\nocite{blom2015click}
\nocite{chen2015misleading}
\nocite{ecker:2014effects}
\nocite{chakraborty2016stop}
\end{document}